\documentclass{article}
\usepackage{spconf,amsmath,graphicx}

\usepackage{enumitem}
\usepackage{tabularx,booktabs, comment}
\usepackage{url}
\usepackage{comment}
\usepackage{multirow}
\setlist{nosep, leftmargin=14pt}
\usepackage{amsmath}
 \usepackage{colortbl}

\usepackage{mwe} 

\definecolor{greenEQ}{RGB}{154,255,153} 
\definecolor{redNEQ}{RGB}{255,204,201}  

\title{MRI-to-CT Synthesis with Cranial Suture Segmentations using a Variational Autoencoder Framework}
\name{\begin{tabular}{@{}c@{}}
Krithika Iyer$^{\ast1}$, Austin Tapp$^{\ast1}$, Athelia Paulli$^{1,2,3}$, Gabrielle Dickerson$^{1}$, Syed Muhammad Anwar$^{1,3}$,\\ 
Natasha Lepore$^{2}$, and Marius George Linguraru$^{1,3}$\thanks{$\ast$ Equal contribution.}\vspace{-5mm}
\end{tabular}}

\address{\begin{tabular}{@{}p{0.96\textwidth}@{}}
\centering
$^{1}$ Sheikh Zayed Institute for Pediatric Surgical Innovation, Children’s National Hospital, Washington, DC, USA\\
$^{2}$ CIBORG Lab, Department of Radiology, Children’s Hospital Los Angeles, Los Angeles, CA, USA
$^{3}$ Departments of Radiology and Biomedical Engineering, University of Southern California, Los Angeles, CA, USA\\
$^{4}$ School of Medicine and Health Sciences, George Washington University, Washington, DC, USA
\end{tabular}\vspace{-5mm}}

\begin{document}
\sloppy
%
\maketitle

\begin{abstract}
\raggedbottom
Quantifying normative pediatric cranial development and suture ossification is crucial for diagnosing and treating growth-related cephalic disorders. Computed tomography (CT) is widely used to evaluate cranial and sutural deformities; however, its ionizing radiation is contraindicated in children without significant abnormalities. Magnetic resonance imaging (MRI) offers radiation-free scans with superior soft-tissue contrast, but unlike CT, MRI cannot elucidate cranial sutures, estimate skull bone density, or assess cranial vault growth.
This study proposes a deep learning-driven pipeline for transforming T1-weighted MRIs of children aged 0.2–2 years into synthetic CTs (sCTs), predicting detailed cranial bone segmentation, generating suture-probability heatmaps, and deriving direct suture segmentation from the heatmaps.
With our in-house pediatric data, sCTs achieved 99\% structural similarity and a Fréchet inception distance of 1.01 relative to real CTs. Skull segmentation attained an average Dice coefficient of 85\% across seven cranial bones, and sutures achieved 80\% Dice. Equivalence of skull and suture segmentation between sCTs and real CTs was confirmed using two one-sided tests (TOST; $p<0.05$). To our knowledge, this is the first pediatric cranial CT synthesis framework to enable suture segmentation on sCTs derived from MRI, despite MRI’s limited depiction of bone and sutures. By combining robust, domain-specific variational autoencoders, our method generates perceptually indistinguishable cranial sCTs from routine pediatric MRIs, bridging critical gaps in non-invasive cranial evaluation.
\end{abstract}
\begin{keywords}
pediatric neuroimaging, cranial sutures, cranial growth, MRI, synthetic CT
\end{keywords}

\sloppy
\section{Introduction}
In early childhood, brain and cranial development are interconnected, with the expanding cranium accommodating rapid volumetric and structural growth of the brain. The synchronized maturation is driven by cranial sutures (fibrous joints connecting the eight skull bones), which remain patent to permit vault expansion during the first years of life before gradually ossifying~\cite{C18,C21}. Sutures function both as growth sites and as flexible boundaries essential for maintaining normal head shape and accommodating intracranial growth. Premature suture fusion, as seen in cephalic disorders, can disrupt neuro-developmental and craniofacial symmetry \cite{C23}. Timely characterization of normative morphological suture progression is clinically important, given the prevalence of cranial deformities in infants~\cite{C19,C20}. Despite their central role in cranial morphogenesis, sutures and their gradual ossification remain poorly represented in infant growth and segmentation models~\cite{C21}.

Clinical assessment of sutures and cranial bones traditionally relies on computed tomography (CT), which provides excellent contrast for osseous structures and direct depiction of suture morphology~\cite{falconi2023imaging}. However, CT’s ionizing radiation is contraindicated for routine screening in children who may not present with overt pathology, limiting its utility in longitudinal or population studies \cite{almohiy2014paediatric}. Magnetic resonance imaging (MRI) is preferred for pediatric neuroimaging because it avoids the use of ionizing radiation and offers excellent soft-tissue contrast. However, MRI does not directly visualize cranial bone or suture anatomy ~\cite{sreedher2021cranial}. MRI inherently has low bone contrast. Consequently, even optimally windowed MRIs often fail to accurately depict thin pediatric skull sutures, limiting their clinical utility for suture assessment ~\cite{sreedher2021cranial}.

Recent advances in deep learning generative models have improved pediatric cranial MRI-to-CT synthesis ~\cite{tapp2025mri, tappISBI}. Yet, explicit pediatric suture analysis remains largely unaddressed, which limits non-invasive growth assessment using routine MRI. Most MRI-to-CT synthesis approaches utilize Generative Adversarial Networks (GANs) ~\cite{GAN2014}, Variational Autoencoders (VAEs) ~\cite{kingma2013auto}, or diffusion models \cite{ho2020denoising}. While GANs achieve high perceptual quality, they often introduce inter-slice discontinuities, lose thin structures, and may hallucinate anatomy~\cite{C9}. VAE-based models offer stable training and compact representations but tend to over-smooth fine osseous details like sutures~\cite{C2,C12}. Diffusion-based approaches improve anatomical fidelity ~\cite{C12} but are computationally intensive in 3D~\cite{C12}; while latent diffusion is computationally feasible, it often lacks explicit anatomical constraints, risking misalignment of thin structures~\cite{tappISBI,C12}. A synthesis strategy that considers global cranial structure to recover explicit, thin anatomical features is therefore necessary to enable the accurate localization of sutures and quantify suture morphology, including, but not limited to, their patency, position, and size.

To this end, we propose a novel dual-stage VAE framework for parallel MRI-to-CT synthesis, cranial bone segmentation, and probabilistic suture segmentation. Our approach harnesses two VAEs: (i) the first VAE synthesizes synthetic CT (sCT) from MRI, preserving clinically relevant skull morphology and structure using adversarial training to maximize sCT realism,  (ii) the second VAE leverages sCT outputs from (i) and an anatomical atlas to segment 7 cranial bones (bilateral frontal, bilateral temporal, bilateral parietal, and occipital) as well as sutures. By harnessing the probabilistic nature of atlas-informed VAE segmentation, our pipeline also generates interpretable probability heatmaps that convey the confidence of model predictions, a crucial feature for thin structures like sutures, ensuring usability in clinical applications. The atlas-based anatomical prior mitigates the lack of explicit suture signal in MRI-derived sCTs, providing morphological context that steers the segmentation branch toward anatomically plausible suture predictions. 
The proposed pipeline offers a radiation-free, anatomically informed solution for pediatric cranial evaluation, including suture analysis, thereby supporting longitudinal and population studies essential for early diagnosis and management of cranial diseases and disorders.

\sloppy
\raggedbottom
\section{Methods}

Paired MRI and CT scans of healthy pediatric patients were retrospectively collected from Children’s National Hospital (Washington, DC, USA) as part of standard clinical care. The cohort included infants whose MRI and CTs were acquired within 30 days of each other, resulting in 116 subjects (64 male, 52 female) aged 0.12–2 years (mean 0.85 years). CT scans had an average in-plane pixel spacing of 0.39\,mm and an average slice thickness of 1.02\,mm. Corresponding T1-weighted MRIs acquired at 1.5\,T or 3\,T had an average in-plane pixel spacing of 0.57\,mm and an average slice thickness of 0.44\,mm. All images were resized to $224\times224\times224$~mm$^{3}$. 
All CT scans underwent bed removal to isolate the head and suppress scanner-associated artifacts, followed by rigid 6-degree-of-freedom (6-DOF) alignment to a reference CT. 
MRI volumes were corrected for intensity nonuniformity (bias field) and then registered to their corresponding CT scans using a 9-degree-of-freedom (rotation, translation, and scaling) transform to ensure cross-modality spatial correspondence.

Cranial bone segmentations were obtained using the publicly available model by Liu et al.~\cite{liu2023joint}. This method uses a DenseNet-inspired network with a context-encoding module that jointly performs bone labeling and cranial-base landmark localization. The model was trained on 274 normative subjects (147 male, aged $0.85\pm0.57$ years) manually annotated by expert raters who labeled the calvaria. On held-out testing cohorts, the model achieved an average Dice similarity coefficient (DSC) of 91\% on normative subjects. Using this model, we obtained 7 skull bone segmentations for our cohort of 116 paired CTs, manually reviewed the predictions, and used them as the ground-truth segmentations.

Cranial sutures were manually segmented by experts from an independent set of 24 unpaired CT scans ($0.97\pm0.72$, 12 male), separate from the MRI–CT paired cohort. These segmentations served as ground truths for training an nnU-Net to automatically annotate cranial sutures~\cite{C25}. The nnU-Net annotation model was iteratively refined using human-in-the-loop retraining until 5-fold cross-validation convergence at 80\% DSC for direct suture segmentations. The final nnU-Net model annotated our dataset of 116 paired CTs which were manually inspected for accuracy and served as ground-truth suture labels.

The youngest patient's scan was selected as the reference atlas and reference CT (for rigid alignment) based on two primary criteria. The scan exhibited excellent quality, with complete cranial structure, clean tissue contrast, and minimal artifacts, thus ensuring anatomical fidelity for downstream alignment. Second, younger subjects feature an open fontanelle and wider cranial sutures, providing optimal visibility and anatomical landmarks. This facilitates robust supervision of suture localization during network training, leveraging the clear depiction of fontanelles and suture positions present in early infancy.
\subsection{Proposed Model}\label{proposed_model}
\begin{figure}[!ht]
    \centering
    \includegraphics[width=\linewidth]{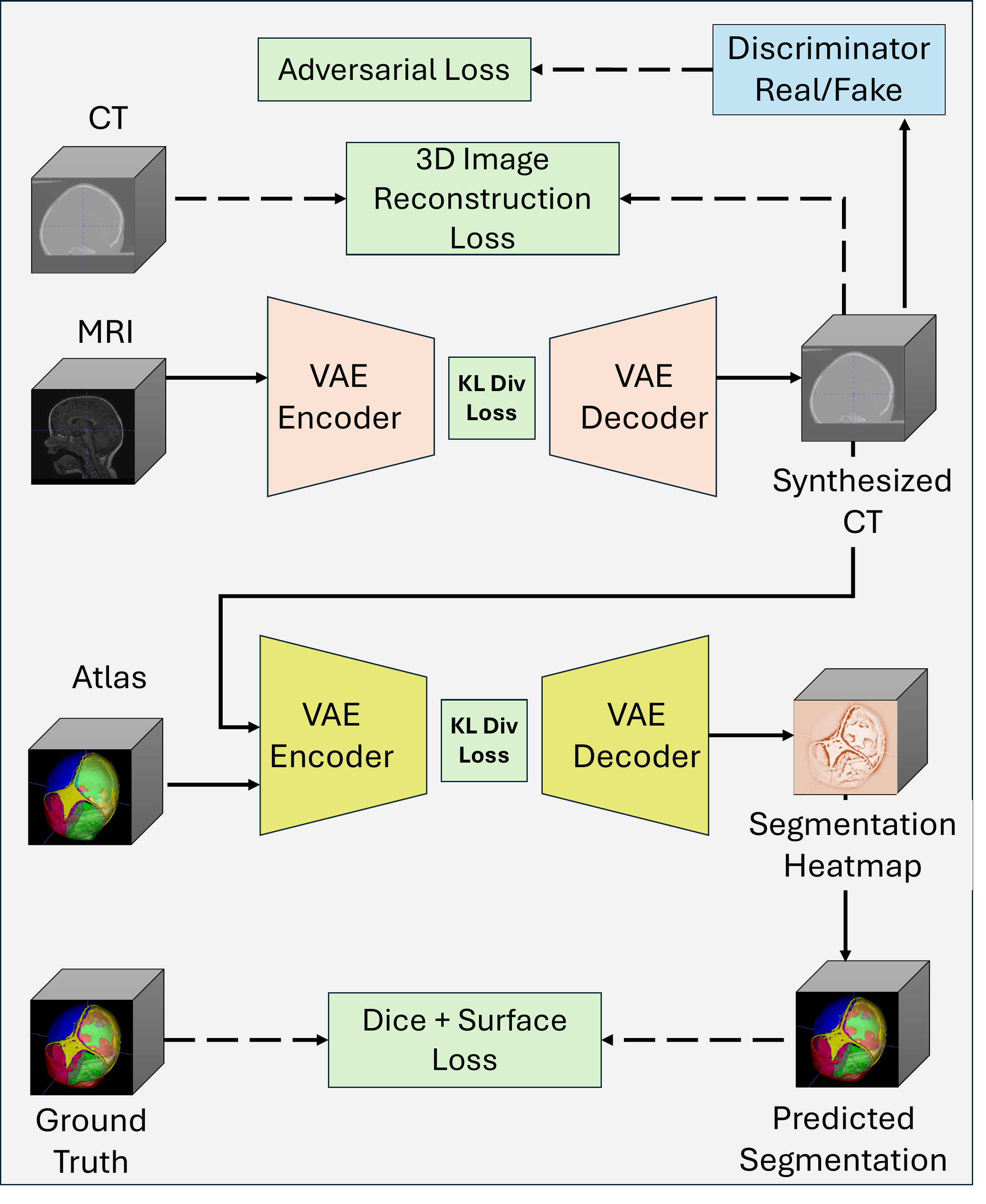}
    \vspace{-8mm}
    \caption{Proposed framework: A VAE generates synthetic CT from input MRI, optimized with adversarial and 3D reconstruction losses. The synthesized CT is processed by a VAE-based, atlas-informed segmentation module that outputs probabilistic segmentation labels.}
    \label{fig:block_diagram}
\end{figure}

Our framework uses two coordinated components: MRI-to-CT synthesis and atlas-guided segmentation, leveraging the strengths of generative models and anatomical conditioning to address pediatric cranial imaging challenges.
\begin{enumerate}
    \item \textbf{MRI-to-CT Synthesis:} We employ a Medical AI for Synthetic Imaging (MAISI)-based \cite{guo2025maisi} VAE generative framework, which has demonstrated robust generalization across diverse anatomical regions and clinical conditions. The encoder compresses input T1-weighted pediatric MRIs into a latent space capturing global cranial features. The decoder reconstructs sCTs, and model training minimizes a composite loss consisting of voxel-wise reconstruction loss (preserving structural and fine-grained detail), adversarial loss (promoting realistic image generation), Kullback-Leibler (KL) divergence (ensuring latent space regularity), and perpetual realism loss. This multi-component approach ensures that the sCTs faithfully match the anatomical and structural properties of paired MRIs, while maintaining realistic tissue contrast, appearance, and fine bone detail essential for downstream segmentation and analysis.
    \item \textbf{Atlas-Guided Segmentation}: In the subsequent branch, the sCTs are concatenated with our segmentation atlas that contains seven cranial bone labels and the suture label. This two-channel input explicitly provides morphological context for segmentation. The composite volume is processed by a second autoencoder network, structured analogously to the MAISI backbone but omitting the adversarial discriminator for this stage. The segmentation network predicts an eight-label output. Training objective uses a combination of Dice-Focal loss and Hausdorff distance loss. \\
\end{enumerate}
By coupling CT synthesis with atlas-driven segmentation, our model learns to localize sutures from global cranial morphology as represented by both the sCTs and the atlas template, even where explicit suture features are absent, as is the case for MRI-derived sCTs. Probabilistic heatmaps for suture segmentation further support model interpretability and confidence estimation, a critical requirement for clinical deployment.

\sloppy
\begin{figure}[th]
    \centering
    \includegraphics[width=\linewidth]{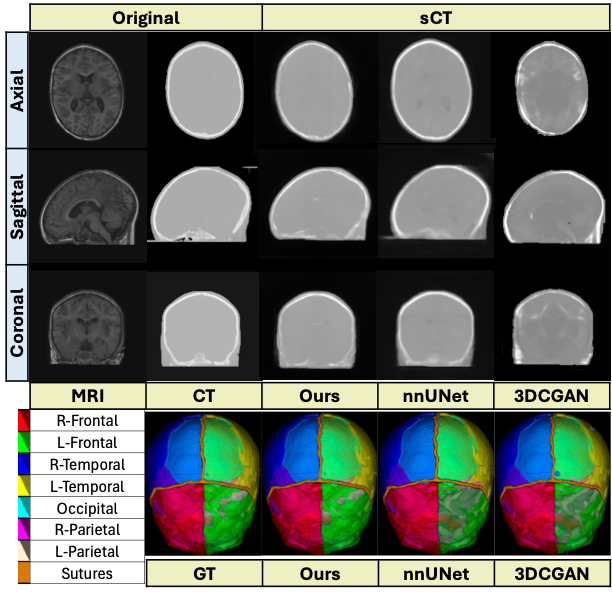}
    \vspace{-8mm}
    \caption{Top: Axial, sagittal, and coronal slices for MRI, ground truth CT, and synthesized CT (sCT) using different methods (Ours, nnUNet, 3DCGAN). Bottom: 3D ground truth and predicted cranial bone and suture segmentations for each method. Age = 683 days, Sex = Female}
    \label{fig:qualitative_synthesis}
\end{figure}
\begin{figure}[ht]
    \centering
    \includegraphics[width=1\linewidth]{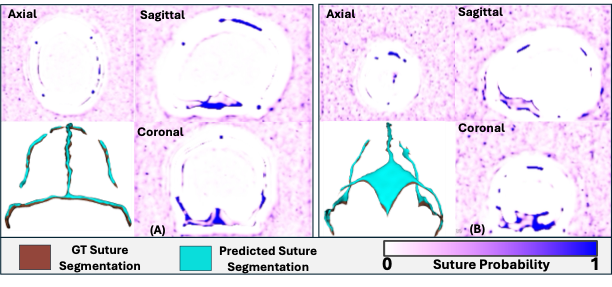}
    \vspace{-8mm}
    \caption{Axial, sagittal, and coronal views show suture probability heatmaps along with 3D visualization of GT suture segmentation (brown) and predicted segmentation (cyan) overlaid for two samples. (A) Age = 683 days, Sex = Female (B) Age = 80 days, Sex = Female}
    \label{fig:prob_suture}
\end{figure}

\begin{table}[!]
\centering
\renewcommand{\arraystretch}{1}
\setlength{\tabcolsep}{4pt}
\scriptsize
\caption{CT image quality metrics comparison across 3DCGAN, nnU-Net, and our model.}
\begin{tabular}{lccc}
\toprule
\textbf{Metric} & \textbf{3DCGAN} & \textbf{nnU-Net} & \textbf{Ours} \\
\midrule
FID $\downarrow$   & 3.99 ± 2.23  & 1.64 ± 0.98 & \textbf{1.02 ± 0.45} \\
SSIM $\uparrow$    & 0.97 ± 0.01  & \textbf{0.99 ± 0.01}  & \textbf{0.99 ± 0.00} \\
LPIPS $\downarrow$ & 0.04 ± 0.02  & \textbf{0.01 ± 0.01} & \textbf{0.01 ± 0.01} \\
PSNR $\uparrow$    & 26.06 ± 3.57 & 31.98 ± 3.68 & \textbf{33.68 ± 3.38} \\
MAE $\downarrow$   & 0.57 ± 0.10  & 0.32 ± 0.16  & \textbf{0.24 ± 0.09} \\
\bottomrule
\end{tabular}
\vspace{-4mm}
\label{tab:ct_metrics}
\end{table}

\begin{table}[!]
\centering
\renewcommand{\arraystretch}{1}
\setlength{\tabcolsep}{3pt}
\tiny
\caption{Segmentation performance comparison across anatomical regions for 3DCGAN, nnU-Net, and our models.}
\begin{tabular}{lcccccc}
\toprule
\textbf{Label} & \textbf{3DCGAN} & \textbf{3DCGAN (FT)} & \textbf{nnU-Net} & \textbf{nnU-Net (FT)} & \textbf{Ours} & \textbf{Ours (FT)} \\
\midrule
R-Frontal    & 0.78 ± 0.06 & 0.75 ± 0.09 & 0.80 ± 0.04 & 0.81 ± 0.04 & 0.82 ± 0.04 & \textbf{0.83 ± 0.06} \\
L-Frontal    & 0.84 ± 0.05 & 0.75 ± 0.13 & 0.86 ± 0.03 & 0.83 ± 0.05 & \textbf{0.88 ± 0.03} & 0.86 ± 0.06 \\
R-Temporal   & 0.91 ± 0.02 & 0.92 ± 0.02 & 0.91 ± 0.02 & 0.92 ± 0.02 & 0.92 ± 0.02 & \textbf{0.92 ± 0.02} \\
L-Temporal   & 0.91 ± 0.02 & 0.93 ± 0.02 & 0.91 ± 0.03 & 0.93 ± 0.02 & 0.91 ± 0.02 & \textbf{0.93 ± 0.02} \\
Occipital    & 0.87 ± 0.04 & 0.87 ± 0.06 & 0.90 ± 0.03 & 0.89 ± 0.05 & \textbf{0.91 ± 0.02} & 0.90 ± 0.03 \\
R-Parietal   & 0.84 ± 0.07 & 0.69 ± 0.05 & 0.85 ± 0.07 & 0.70 ± 0.04 & \textbf{0.92 ± 0.02} & 0.83 ± 0.09 \\
L-Parietal   & 0.81 ± 0.11 & 0.46 ± 0.18 & 0.82 ± 0.12 & 0.46 ± 0.16 & \textbf{0.92 ± 0.03} & 0.76 ± 0.19 \\
Sutures      & 0.78 ± 0.06 & 0.80 ± 0.06 & 0.78 ± 0.06 & 0.80 ± 0.06 & 0.79 ± 0.06 & \textbf{0.81 ± 0.06} \\
\midrule
Mean Dice    & 0.84 ± 0.05 & 0.77 ± 0.07 & 0.85 ± 0.05 & 0.79 ± 0.05 & \textbf{0.88 ± 0.03} & 0.85 ± 0.06 \\
Mean HD95    & 20.34 ± 12.16 & 28.17 ± 12.05 & 11.23 ± 7.53 & 20.19 ± 8.77 & \textbf{7.86 ± 5.20} & 13.98 ± 15.12 \\
\bottomrule
\end{tabular}
\label{tab:segmentation_comparison}
\end{table}

\section{Results}

We adopted the MAISI autoencoder architecture with its default parameters \cite{guo2025maisi}. Our model was compared against two state-of-the-art (SOTA) MRI-to-CT synthesis networks: (1) 3D CycleGAN \cite{ge2019} and (2) nnU-Net \cite{Arthur2024}. All baseline models were trained with their default hyperparameters, using a consistent protocol of 1000 training epochs and an age- and sex-balanced 80/10/10 data split.
For fair segmentation comparison, our segmentation module was applied to the sCTs synthesized from each respective MRI-to-CT synthesis model. Additionally, we present two segmentation results for each model: (1) using the proposed network described in Section~\ref{proposed_model}.2, and (2) using the same proposed network but fine-tuned (FT) to generalize to sCTs. Thus, we report both standard and fine-tuned results for all methods. 

Fig.~\ref{fig:qualitative_synthesis} demonstrates that our pipeline generates realistic, high-quality sCTs and corresponding segmentations, outperforming the comparison models. 
Fig.~\ref{fig:prob_suture} shows suture-probability heatmaps and 3D renderings of predicted sutures overlaid on ground-truth skull segmentations for two representative test cases: (a) an older infant (683\,days) and (b) a younger infant (80\,days) with a patent fontanelle. Across both cases, the model captures the expected sutural courses and relative patency, with the heatmaps reflecting high probabilities along sutural margins. Notably, initializing the atlas from the youngest subject does not degrade performance on older infants as accurate suture localization and bone segmentations are preserved across the evaluated pediatric age range (Fig.~\ref{fig:qualitative_synthesis}, Fig.~\ref{fig:prob_suture}). These results support the feasibility of radiation-free, MRI-based assessment of pediatric cranial development.

We evaluated perceptual quality of sCTs using Fréchet Inception Distance (FID), structural similarity (SSIM), learned perceptual image patch similarity (LPIPS), peak signal-to-noise ratio (PSNR), and mean absolute error (MAE). Segmentation accuracy was measured with the DSC and the 95th-percentile Hausdorff distance (HD95). Summary results are reported in Table~\ref{tab:ct_metrics} (image similarity) and Table~\ref{tab:segmentation_comparison} (segmentation). Our framework demonstrates SOTA performance for perceptual image quality and segmentations; it consistently surpasses baselines in FID, PSNR, and MAE while attaining high Dice for cranial bones (88\%) and sutures (79\%), despite their thin, low-contrast nature. These findings demonstrate that the method produces high-fidelity sCTs and accurate automated cranial segmentations directly from T1-weighted MRI.

To evaluate the statistical significance of the differences in the performances of the models, we compared paired segmentation performance against the baselines (3DCGAN and nnU\mbox{-}Net) using the Wilcoxon signed-rank test. Differences versus 3DCGAN were significant (\(p<0.05\)) in all regions except the left temporal, and versus nnU\mbox{-}Net in all regions except the right temporal and occipital, indicating consistent improvements as a result of preserving relevant morphology.

We also assessed equivalence between real CT and sCT segmentations using the two one-sided tests (TOST) procedure with prespecified bounds (Dice \( \pm 0.02\), HD95 \( \pm 3\,\mathrm{mm}\)). Our method achieved statistical equivalence (\(p<0.05\)) in all regions except the parietal bones, whereas the baselines failed to meet equivalence in 5 of 8 regions. Together with superior DSC and HD95, these results support that our approach produces sCTs with greater anatomical fidelity and realism.

This study has several limitations. Suture supervision originated from 24 manually labeled CTs and was expanded through pseudo-labeling. In addition, the cohort is single-center and limited to infants aged 0.12–2 years, which may affect generalizability to older age ranges and syndromic presentations. Including clinical validation (such as assessing neurodevelopmental outcomes' relationship to morphological features) will further strengthen the utility of the framework. As data scale increases, exploring atlas-free conditioning may also help reduce potential biases introduced by atlas selection.

\sloppy
\section{Conclusion}
We introduced a dual-stage VAE pipeline that converts routine pediatric T1-weighted MRI into sCT, performs atlas-guided cranial bone segmentation, and localizes sutures. 
Our approach achieved significantly better paired performance in most regions compared to SOTA models (Wilcoxon signed-rank; \(p<0.05\)) and reached statistical equivalence to real CT (TOST; \(p<0.05\)) for all but the parietal bones. Qualitative results further demonstrate that the predicted follow anatomically plausible sutural courses across age.
Our approach represents a significant advance toward reliable, radiation-free evaluation of cranial bones and sutures in early life, with potential to support longitudinal and population studies for early diagnosis and management of cranial disorders.



\clearpage
\section{Acknowledgments}
This research study was conducted retrospectively using human subject data made available by Children's National Hospital, Washington, DC, USA. Approval was granted by the Ethics Committee of the Children's National Hospital. 
This work was supported by NIH NIDCR R01DE030286.
The authors have no relevant financial or non-financial interests to disclose.
\vspace{-5mm}
\sloppy
\bibliographystyle{IEEEbib}
\bibliography{strings,refs}

\end{document}